\def\year{2020}\relax
\documentclass[letterpaper]{article} 
\usepackage{aaai21}  
\usepackage{times}  
\usepackage{helvet} 
\usepackage{courier}  
\usepackage[hyphens]{url}  
\usepackage{graphicx} 
\usepackage{natbib}  
\usepackage{caption} 
\usepackage{subcaption}
\usepackage[utf8]{inputenc} 
\usepackage[T1]{fontenc}    
\usepackage{url}            
\usepackage{booktabs}       
\usepackage{amsfonts}       
\usepackage{nicefrac}       
\usepackage{microtype}      
\usepackage{xcolor}
\usepackage{amsmath}
\usepackage{mathtools}
\usepackage{amsthm}
\usepackage{graphicx}
\usepackage{caption}
\usepackage{lineno}
\usepackage{comment}
\usepackage{cite}
\usepackage{amsmath,amssymb,amsfonts}
\usepackage{comment}
\usepackage{float}
\usepackage{multirow}
\usepackage{makecell}
\usepackage{caption}
\usepackage{subcaption}
\usepackage{graphicx}
\usepackage{textcomp}
\usepackage{xcolor}
\usepackage{setspace}
\usepackage{algorithmic}
\usepackage[ruled,linesnumbered]{algorithm2e}
\usepackage{pifont}
\usepackage{amssymb}
\usepackage{pifont}
\newcommand{\xmark}{\ding{55}}%

\usepackage{bbding}

\DeclarePairedDelimiterX\set[1]\lbrace\rbrace{#1}

\newcommand{\argmin}{\operatornamewithlimits{argmin}}

\newcommand*{\email}[1]{\texttt{#1}}
\newcommand\todo[1]{\textcolor{red}{#1}}

\newcommand*\Mname{\textcolor{black}{SAG}}
\usepackage[flushleft]{threeparttable}

\usepackage{mathtools}

\DeclarePairedDelimiter\floor{\lfloor}{\rfloor}

\title{Scalable Adversarial Attack on Graph Neural Networks \\ with Alternating Direction Method of Multipliers}
\author{
    Boyuan Feng, Yuke Wang, Xu Li, and Yufei Ding \\
}

\affiliations{
    University of California, Santa Barbara, USA \\
    \email{boyuan@cs.ucsb.edu, yuke\_wang@cs.ucsb.edu} \\ 
    \email{lixu9906@gmail.com, yufeiding@cs.ucsb.edu}\\
}

\begin{document}


\maketitle

\begin{abstract}
Graph neural networks (GNNs) have achieved high performance in analyzing graph-structured data and have been widely deployed in safety-critical areas, such as finance and autonomous driving.
However, only a few works have explored GNNs' robustness to adversarial attacks, and their designs are usually limited by the scale of input datasets (\textit{i.e.}, focusing on small graphs with only thousands of nodes).
In this work, we propose, SAG, the first scalable adversarial attack method with Alternating Direction Method of Multipliers (ADMM).
We first decouple the large-scale graph into several smaller graph partitions and cast the original problem into several subproblems.
Then, we propose to solve these subproblems using projected gradient descent on both the graph topology and the node features that lead to considerably lower memory consumption compared to the conventional attack methods.
Rigorous experiments further demonstrate that SAG can significantly reduce the computation and memory overhead compared with the state-of-the-art approach, making SAG applicable towards graphs with large size of nodes and edges.
\end{abstract}

\section{Introduction} \label{sec:intro}


Graph neural networks (GNNs) play an essential role in many emerging applications with graph data.
On these applications, GNNs show their strength in extracting valuable information from both the \textit{features} (\textit{i.e.}, information from individual nodes) and the \textit{topology} (\textit{i.e.}, the relationship between nodes).
For example, GNNs can effectively analyze financial data to decide loan-related policies.
Another example is the social network with billions of users where GNNs can do friendship recommendation.

Such wide deployment of GNNs motivates the investigation of their robustness and reliability. 
One of the key aspects is to effectively generate adversarial examples on graph data, so that we can better understand the ``weakness'' of GNNs and secure them more wisely afterward.
Our initial exploration and studies identify several key properties to be considered in the GNN attack.
First, the adversarial example needs to consider \textit{topology} and \textit{feature} information to comprehensively attack the GNNs on all perspectives.
Second, the attack method needs to be efficient in both \textit{memory} and \textit{computation} for catering to the huge number of nodes in graph data.


However, existing work inevitably falls in short at least one of the above aspects, as summarized in Table~\ref{tab:importantProperties}. 
Specifically, FGSM~\cite{FGSM} is crafted for attacking traditional deep neural networks (DNNs). 
Even though it can attack node embeddings with good computation and memory efficiency, it can not well support the graph topology, which distinguishes GNNs from DNNs.
PGD~\cite{xu2019topology} is one of the state-of-the-art work designated for the GNN attack. However, it does not support the attack on node features, and it suffers a quadratic memory overhead due to maintaining the large size of edge gradients in a dense matrix format. 
For example, a graph with $N$ nodes leads to a dense $N\times N$ gradient matrix, consuming more than $10$GB memory for a moderate-size graph with only $50,000$ nodes.
Another work, Nettack~\cite{zugner2018adversarial}, while performing well in three aspects, finds its shortcoming in computation efficiency due to a fine-grained edge manipulation in a very inefficient trial-and-error manner.

\begin{table}[t] \small
    \centering
    \caption{Comparison with the Existing Attack Methods.}
    \vspace{-5pt}
    \scalebox{0.9}{
    \begin{tabular}{|c|c c c c|}
    \hline
        \textbf{Method} & \textbf{Topology} & \textbf{Feature} & \textbf{Comp. Effi.} & \textbf{Mem. Effi.}\\
    \Xhline{2\arrayrulewidth}
        FGSM & \xmark & \Checkmark & \Checkmark & \Checkmark \\
        PGD & \Checkmark & \xmark & \Checkmark & \xmark \\
        Nettack & \Checkmark & \Checkmark & \xmark & \Checkmark \\
    \textbf{\Mname} & \Checkmark & \Checkmark & \Checkmark & \Checkmark \\
    \hline
    \end{tabular}}
    \vspace{-10pt}
    \label{tab:importantProperties}
    \vspace{-5pt}
\end{table}

To overcome these challenges, we propose \Mname, the first ADMM-attack on GNNs with the alternating direction method of multipliers (ADMM), which iteratively maximizes the training loss through modifying the graph topology and the node features. 
ADMM has been shown to be effective in dividing-and-conquering a large non-convex optimization problem into multiple smaller ones for achieving both efficiency and scalability \cite{ijcai2017-228,DBLP:conf/aaai/LengDLZJ18,ADMM-AutoML,ADMM-NN}.
As shown in the last line of Table~\ref{tab:importantProperties}, \Mname~can attack both the graph topology and the node features, while maintaining computation and memory efficiency to a great extent, making it a viable and promising solution for large-scale graphs. 

In summary, our major contributions are:
\begin{itemize}
    \item We identify and analyze the key properties of the effective adversarial attacks on GNNs, where none of the existing methods could address all of them systematically and comprehensively.
    \item We propose \Mname~to effectively generate adversarial examples on graph neural networks based on both topology attack and feature attack.
    We formulate \Mname~with the ADMM optimization framework and achieve both high memory and computation efficiency.
    \item Evaluation shows our proposed method can launch more effective attacks compared with the state-of-the-art approaches while reducing the computation and memory overhead, making it applicable towards large graph settings.
\end{itemize}

\section{Related Work} \label{sec:related_work}

\textbf{Graph Adversarial Attacks.}
Graph adversarial attacks aim to maximize the accuracy drop on GNN models by introducing the perturbations, such as the modifications of the graph topology and the node representations (feature embeddings).
Existing GNN attacks can be broadly classified into two major categories, \textit{poisoning} attack~\cite{zugner2018adversarial,zugner2019adversarial} and \textit{evasion}~\cite{dai2018adversarial} attack, depending on the time they happen. The former (poisoning attack) happens during the training time of the GNNs through modifying training data, while the latter (evasion attack) takes place during the GNN inference time by changing test data samples. Our proposed \Mname~is a comprehensive attack method that can cover both types of attack meanwhile offering significant computation and memory efficiency compared with the existing attack approaches.

\vspace{4pt}
\noindent \textbf{ADMM Framework.}
ADMM \cite{ADMM} is an optimization framework that is effective in decomposing and solving optimization problems under constraints.
The theoretical results of ADMM have been explored in \cite{ADMM_theory1, ADMM_theory2, ADMM_theory3,ADMM_Theory4} for various convex problems under diverse constraints and are shown to have linear convergence.
Formally, ADMM can effectively solve an optimization problem under linear constraints
\begin{equation} \small
\begin{split}
    \min_{\mathbf{x},\mathbf{z}} \;\;\;\; & f(\mathbf{x}) + g(\mathbf{z}) \\
    \text{subject to}        \;\;\;\; & A\mathbf{x}+B\mathbf{z} = \mathbf{c}
\end{split}
\end{equation}
where $f(\mathbf{x})$ and $g(\mathbf{z})$ could be either differentiable or non-differentiable but has some exploitable structure properties.
Then, by introducing the augmented Lagrangian function, ADMM can break the problem into two subproblems in $\mathbf{x}$ and $\mathbf{z}$ and iteratively solve each subproblem at one time.
While popular stochastic gradient descent (SGD) method can usually solve optimization problems, SGD cannot effectively process the case with diverse constraints (\textit{e.g.}, equality constraints between variables) and usually require ad-hoc modifications on gradients.

Although ADMM is originally developed to solve convex problems, recent studies successfully exploit ADMM to solve NP-hard non-convex problems under constraints in CNN pruning \cite{ijcai2017-228,ADMM-Pruning}, compressive sensing \cite{NIPS2016_6406}, Auto-ML \cite{ADMM-AutoML}, Top-K feature selection \cite{ijcai2017-228}, and hardware design \cite{ADMM-NN}.
In this paper, we focus on exploring the benefit of exploiting ADMM framework in the context of graph-based adversarial attacks.

\begin{table}[t] \small
    \centering
    \caption{Notations and Definitions used in our paper.}
    \scalebox{0.9}{
    \begin{tabular}{c|l}
    \hline
    \hline
        $G$ & A single large graph \\
        $N$ & The number of nodes \\
        $D$ & The length of node features \\
        $E$ & The number of edges in the original input graph \\
        $A$ & The original input graph of shape $N\times N$ \\
        $X$ & The original input feature of shape $N\times D$ \\
        $W$ & Fixed GNN weights \\
        $S$ & A perturbation matrix of shape $N\times N$  \\
        $\Tilde{A}$ & Perturbed adjacency matrix of shape $N \times N$ \\
        $\Tilde{X}$ & Perturbed features of shape $N\times D$\\
        $\epsilon_A$ & The maximum number of allowed edge perturbations\\
        $\epsilon_X$ & The maximum $L_2$ norm of allowed feature perturbations\\
        $\ell(\cdot,\cdot)$ & The loss function to measure the difference between \\
        & the current output of the model and the targeted labels\\
        $M$ & The number of subproblems to split \\
        $S_i$ & The $i^{th} \in \{1,2,...,K\}$ graph partition of shape $\frac{N}{M} \times N$ \\
        $\Tilde{X}_i$ & The $i^{th} \in \{1,2,...,K\}$ feature copy of shape $N \times D$\\
        $\Pi_{C}[g]$ & The projection of input $g$ towards set $C$ \\
        $K$ & The number of epochs in ADMM \\
        $V_{GT}$ & Set of nodes with ground truth label\\
    \hline
    \hline
    \end{tabular}}
    \label{tab:notation}
    \vspace{-10pt}
\end{table}

\section{Methodology} \label{sec:methodology}

\subsection{Problem Formulation of Scalable Graph Attack}
We first define the notation in this paper, as summarized in Table \ref{tab:notation}.
We consider an input graph $G = (A, X)$, where $A \in \{0,1\}^{N\times N}$ is the adjacency matrix on the edge connection between nodes, $X \in \mathcal{R}^{N\times D}$ is the associated node features, $N$ is the number of nodes in the graph, and $D$ is the feature dimension of each node.
Here, only a portion of nodes $V_{GT}$ are labeled and the goal of node classification is to predict labels for remaining nodes.
Following the common practice in the field \cite{xu2019topology, zugner2018adversarial}, we focus on the well-established work that utilizes graph convolutional layers \cite{GCNConv} for node classification.
Formally, the $i^{th}$ layer is defined as
\begin{equation} \small
    H^{(k+1)} = \Tilde{A}H^{(k)}W^{(k)}
\end{equation}
where $\Tilde{A} = \Tilde{D}^{-\frac{1}{2}}(A+I_N)\Tilde{D}^{-\frac{1}{2}}$ to achieve numerical stability, $D$ is a diagonal matrix with $D_{ii} = \sum_j (A+I_N)_{ij}$, and $W^{(k)}$ is the weights for the $k^{th}$ layer.
Since the memory and the computation complexity generally increase as the number of layers increases, we focus on a single layer GCN that is tractable and still captures the idea of graph convolutions:
\begin{equation} \small
    Z = softmax(\Tilde{A}XW)
\end{equation}
Here, the output $Z \in \{1,2,...,c\}^N$ is the predicted labels for individual nodes.
The parameter $W$ is learned by minimizing the cross-entropy on the output of labeled nodes.
\begin{equation} \small
    \ell(A,X;W, Y_{GT}) = -\sum_{v\in V_{GT}} ln \;\; Z_{v, Y_v}
\end{equation}
where $Y_v$ is the ground truth label for node $v$.
Our experiments show that adversarial examples on this single layer GCN model transfer well to GCNs with various layers and other GNN models such as GAT.
Besides, recent studies \cite{zugner2018adversarial} show that a L-layer GCN can also be treated as $\sigma(A^LXW^L)$ to generate adversarial attacks with tractable computation.

In this paper, SAG aims to generate a perturbed adjacency matrix $\Tilde{A}\in \{0,1\}^{N\times N}$ and a perturbed feature matrix $\Tilde{X} \in \mathcal{R}^{N \times D}$ satisfying a pre-defined perturbation budget.
Similar to \cite{xu2019topology}, we use a Boolean matrix $S \in \{0,1\}^{N\times N}$ to record the edge perturbations that $S_{ij}=1$ indicates the perturbation of edge between node $i$ and node $j$.
Given the original adjacency matrix $A$ and its supplement matrix $\bar{A}$ (\textit{i.e.}, $\bar{A}_{i,j} = \neg A_{i,j}$), we can generate the perturbed adjacency matrix as
$\Tilde{A} = A + (\bar{A} - A) \circ S$,
where $\circ$ is the Hadamard product.
Formally, given the edge perturbation budget $\epsilon_A$ and the feature perturbation budget $\epsilon_X$, \Mname~aims to solve the following optimization problem
\begin{equation} \label{eq:optimization1} \small
\begin{split}
    \min_{S,\Tilde{X}} \;\;\;\;\; & -\ell(S, \Tilde{X}) \\
    \text{s.t.} \;\;\;\;\; & ||\Tilde{X} - X ||_2^2 \leq \epsilon_X \\
                           & 1^TS \leq \epsilon_S, S \in \{0,1\} ^{N\times N}
\end{split}
\end{equation}
For notation simplicity, we use $\ell(S,\Tilde{X})$ to represent the cross-entropy loss $\ell(S, \Tilde{X}; W, Y_{GT})$ when the context is clear.
Following the common practice~\cite{zugner2018adversarial, xu2019topology} in the field, we consider two threat models -- the evasive attack and the poisoning attack.
The evasive attack assumes that the GNN model is fixed and targets the test data.
The poisoning attack targets the training data and performs the model training phase on the perturbed data after the attack.

There are several challenges in solving this optimization problem.
First, the popular stochastic gradient descent (SGD) methods cannot effectively solve optimization problems under constraints and usually require ad-hoc modification on the gradients to satisfying such constraints.
Second, the discrete modification on the edge perturbations makes it a combinatorial optimization problem. \cite{zugner2018adversarial} attacks the graph topology by changing one edge at one time and selecting the edges that achieve the maximum loss.
However, this approach needs to trial-and-error on all edges, leading to prohibitive time cost.
\cite{xu2019topology} takes gradient on the $S$ matrix by releasing the requirement $S \in \{0,1\}^{N \times N}$ to be $S \in [0,1]^{N \times N}$.
However, this approach takes $N\times N$ memory to store the gradient matrix, which is prohibitive for large graphs with tens of thousands of nodes.
Besides, \cite{xu2019topology} only supports topology attack and cannot conduct joint attack on the node features.
We detail ADMM-based SAG on solving the optimization problems under constraints in later sections.

\begin{figure}[t]
    \centering
    \includegraphics[width=\linewidth]{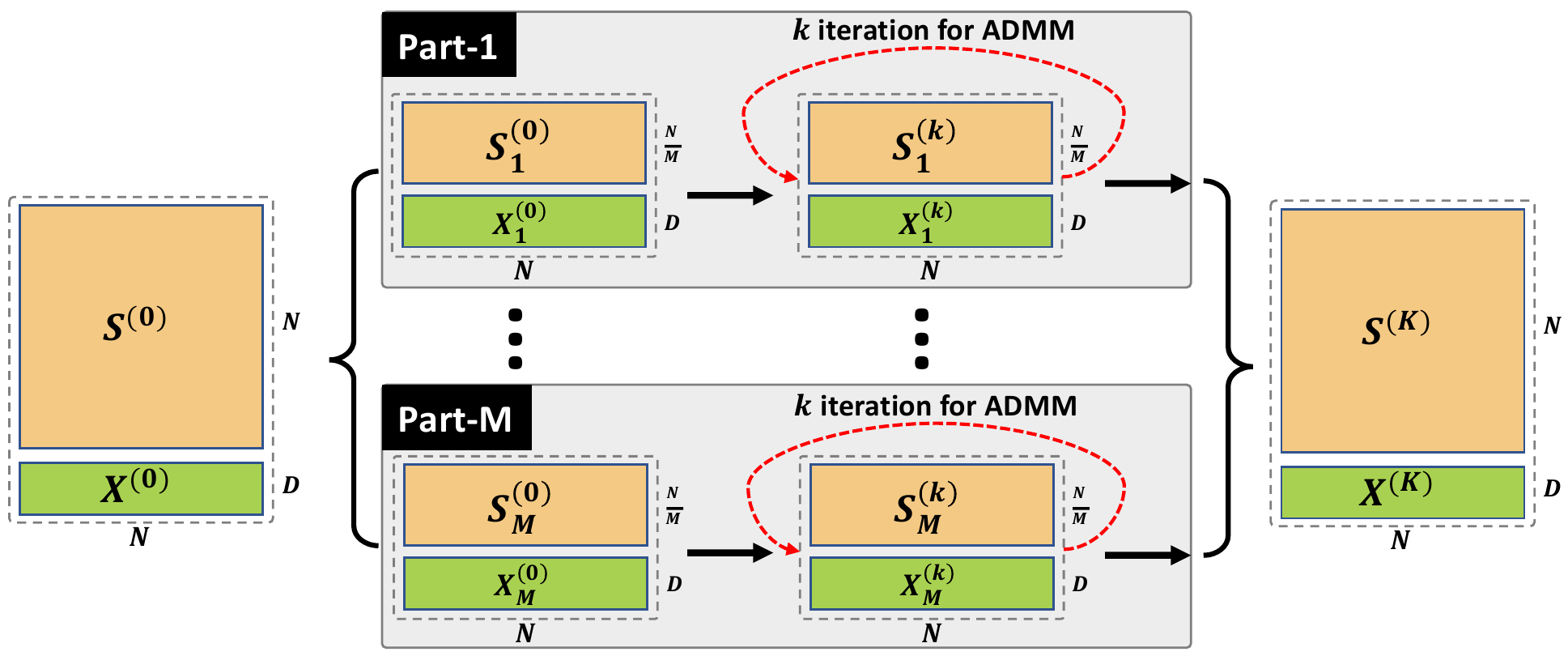}
    \caption{Overview of SAG.}
    \label{fig:overview}
    \vspace{-16pt}
\end{figure}



\subsection{Scalable Graph Attack Framework using ADMM}
SAG achieves a scalable attack on graph data by splitting the matrix $S$ into multiple partitions and consider one partition at each time, as illustrated in Figure~\ref{fig:overview}.
Supposing we split the graph into $M$ partitions, we only need to take gradient on a small matrix with shape $\frac{N}{M} \times N$ that can easily fit into the memory.
Similar to \cite{xu2019topology} and the probabilistic view of node classification problem, we first release the discrete constraints $S \in \{0,1\}^{N\times N}$ to continuous constraints $S \in [0,1]^{N\times N}$, where $S_{ij}$ indicates the probability that an edge needs to be perturbed for attack.
Since the impact between node $i$ and node $j$ may be asymmetry (\textit{i.e.}, the impact of node $i$ on node $j$ is higher than the reverse one), we do not apply symmetry constraint on $S$.
Then, we split $S \in [0,1]^{N \times N}$ into $M$ sub matrix such that $S_i \in [0,1]^{(N/M) \times N}$, where $S_i$ considers the nodes with index between $[\floor*{i*\frac{N}{M}}, \floor*{(i+1)*\frac{N}{M}}]$.
Due to the cumulative property in cross-entropy loss, we can attack by solving the following optimization problem
\begin{equation} \label{eq:optimization2} \small
\begin{split}
    \min_{S_i,\Tilde{X}} \;\;\;\;\; & -\sum_{i=1}^M \ell(S_i, \Tilde{X}) \\
    \text{s.t.} \;\;\;\;\; & ||\Tilde{X} - X ||_2^2 \leq \epsilon_X\\
                           &  1^TS_i \leq \epsilon_{i}, S_i \in [0,1]^{(N/M)\times N}
\end{split}
\end{equation}
Here, $\epsilon_i$ represents the allowed number of edges to change in each graph partition $S_i$ and we set it to be $\frac{\epsilon_A}{M}$ for simplicity.

Ideally, we can split the problem~\ref{eq:optimization2} into $M$ sub-problems and solve them independently for memory efficiency.
However, problem~\ref{eq:optimization2} still has interaction between $M$ sub-problems on the $\Tilde{X}$ term.
To this end, we further reformulate the problem \ref{eq:optimization2} into the following problem by substitute $\Tilde{X} \in \mathcal{R}^{N \times D}$ with a duplicated feature matrix $\Tilde{X}_i \in \mathcal{R}^{N \times D}$
\begin{equation} \label{eq:optimization4} \small
\begin{split}
    \min_{S_i, \Tilde{X}_i} \; & -\sum_{i=1}^M \ell(S_i, \Tilde{X}_i) + \sum_{i=1}^M I_{C_{Si}}(S_i) + \sum_{i=1}^M I_{C_{Xi}}(\Tilde{X}_i)\\
    \text{s.t.} \;\;\;\; & \Tilde{X_i} = \Tilde{X}_{i+1}, \;\;\;\; i\in \{1,2,...,M\} \\
\end{split}
\end{equation}
For notation simplicity, we use $\Tilde{X}_{M+1}$ to represent $\Tilde{X}_1$.
$I_C(x)$ is an indicator function such that $I_C(X) = 0$ if $X\in C$, otherwise $I_C(X) = \infty$.
$C_{Si}$ and $C_{Xi}$ are feasible sets
\begin{equation} \small
\begin{split}
    C_{Si} & = \{S_i  \;| \; S_i \in [0,1]^{\frac{N}{M}\times N}, \;\;1^TS_i \leq \epsilon_{i} \} \\
    C_{Xi} & = \{\Tilde{X_i}  \;| \; ||\Tilde{X_i}-X||_2^2 \leq \epsilon_X  \}\\
\end{split}
\end{equation}

Here, the popular SGD cannot be easily applied to the optimization problem \ref{eq:optimization4} under constraints, especially the equality ones.
To this end, we can adopt the ADMM framework to systematically solve the problem.
First, we have the Lagragian function
\resizebox{.9\linewidth}{!}{
  \begin{minipage}{\linewidth}
\begin{equation} \small
    \begin{split}
        L_\rho(\Tilde{X}_i, S_i, \mu_i) = & -\sum_{i=1}^M\ell(S_i, \Tilde{X}_i) + \sum_{i=1}^M I_{C_{Xi}}(\Tilde{X}_i) + \sum_{i=1}^M I_{C_{Si}}(S_i) \\
        & + \frac{\rho}{2}\sum_{i=1}^M||\Tilde{X}_i - \Tilde{X}_{i+1}||^2 + \sum_{i=1}^M \mu_i^T(\Tilde{X}_i - \Tilde{X}_{i+1})
    \end{split}
\end{equation}
\end{minipage}
}
where $\rho > 0$ is a hyper-parameter and $\mu_i \in \mathcal{R}^{N \times D}$ is the dual variable.

Following the ADMM framework, we can solve the problem \ref{eq:optimization4} by repeating for $k \in \{1,2,...,K\}$ iterations and, in each iteration, solving each $S_i$ and $\Tilde{X}_i$ individually  
\begin{equation} \small \label{eq:ADMMForm}
    \begin{split}
        \Tilde{X}_i^{(k+1)} &= \argmin_{\Tilde{X}_i}L_\rho(\Tilde{X}_i, S_i^{(k)}, \mu_i^{(k)})\\
        S_i^{(k+1)} &= \argmin_{S_i}L_\rho( \Tilde{X}_i^{(k+1)}, S_i, \mu_i^{(k)}) \\
        \mu_i^{(k+1)} &= \mu_i^{(k)} + \rho(\Tilde{X}_i^{(k+1)}-\Tilde{X}_{i+1}^{(k+1)})
    \end{split}
\end{equation}
which is respectively the feature update, topology update, and dual update. 
We stress that we only need to solve the minimization problem for a single graph partition $S_i\in \mathcal{R}^{\frac{N}{M} \times N}$, leading to much reduced memory consumption compared to \cite{xu2019topology}, which requires solving the whole graph $S \in \mathcal{R}^{N \times N}$ at the same time.
Here, the main memory overhead comes from the duplicated feature $\Tilde{X}_i \in \mathcal{R}^{N \times D}$.
However, the feature dimension $D$ is usually a fixed number around $1000$, which is much smaller than the number of nodes $N$ that may reach tens of thousands, or even millions.

\subsection{Algorithm Subroutines for SAG Optimization}
In this section, we present how to efficiently solve the above problem and derive closed form formula for individual minimization problems.

\vspace{3pt}
\noindent \textbf{Feature Update.}
In feature update, we aim to find the feature $\Tilde{X}_i$ that minimizes
\begin{equation} \small
    -\ell(S_i^{(k)}, \Tilde{X}_i) + I_{C_{Xi}}(\Tilde{X}_i) + \frac{\rho}{2} ||\Tilde{X}_i - \Tilde{X}_{i+1}^{(k)}||^2
         + \mu_i^{(k)\;T}(\Tilde{X}_i - \Tilde{X}_{i+1}^{(k)} )
\end{equation}
Here, $\ell(\cdot, \cdot)$ is the cross-entropy loss on the GNN predictions, the last two terms are differentiable functions, but the $I_{C_{Xi}}(\cdot)$ function cannot be easily solved with gradient descent method due to the $\infty$ values for $\Tilde{X}_i \notin C_{Xi}$.
To this end, we adopt a two-step strategy for the feature update.
For the first three terms that are differentiable, we use the gradient descent method to access the gradient $g_{Xi}^{(k)}$ on $\Tilde{X}_i$ and get a pseudo-perturbed feature $\Tilde{X}_i^{(k+1)'}$
\begin{equation} \small
    \Tilde{X}_i^{(k+1)'} = \Tilde{X}_i^{(k)} - \eta_k \cdot g_{X_i}^{(k)}
\end{equation}
where $\eta_k$ is the learning rate at iteration $k$.
Note that the update on feature $\Tilde{X}_i$ depends only on the graph partition $S_i^{(k)}$ and does not involve with remaining graph partitions.
For the term $I_{C_{Xi}}(\cdot)$, we refer to the projection method by projecting the pseudo-perturbed feature onto the feasible set $C_{Xi}$:
\begin{equation} \label{eq:featureUpdate} \small
\begin{split}
    \Tilde{X}_i^{(k+1)} & = \Pi_{C_{Xi}}[\Tilde{X}_i^{(k)} - \eta_k \cdot g_{X_i}^{(k)}] \\
    g_{X_i}^{(k)} & = -\frac{\partial}{\partial X_i} \ell(S_i^{(k)}, X_i^{(k)}) + \rho (X_i^{(k)} - X_{i+1}^{(k)})  + \mu^{(k)}
\end{split}
\end{equation}
While computing the projection is a difficult task in general and usually requires an iterative procedure to solve, we exploit the special structure of $C_{Xi}$ and derive a closed-form formula to analytically solve it.

\vspace{2pt}
\noindent \textbf{Proposition 1}. Given $C_{X} = \{a  \;| \; ||a-X||_2^2 \leq \epsilon_X\}$, the projection of $a$ to $C_X$ is
\begin{equation} \small
    \Pi_{C_{X}}(a) = \begin{cases}
                \frac{a+uX}{1+u} \;\; &\text{if $u>0$ and $u = \sqrt{\frac{||a-X||_2^2}{\epsilon_X}} -1 $} \\
                a \;\; & \text{if $||a-X||_2^2 \leq \epsilon_X$}
            \end{cases}
\end{equation}

\noindent \textbf{Proof:}
$\Pi_{C_X}(a)$ can be viewed as an optimization problem
\begin{equation} \small
    \begin{split}
        \min_R \;\;\;\; & || R - a||_2^2 \\
        \text{s.t.} \;\;\;\; & (R-X)^T(R-X) \leq\epsilon_X
    \end{split}
\end{equation}
We can derive its Lagrangian function as 
\begin{equation} \small
    L(R,u) = || R - a||_2^2 + u[(R-X)^T(R-X) - \epsilon_X]
\end{equation}
where $u \geq 0$.
Using the KKT condition we have the stationary condition that 
\begin{equation} \small
    \frac{\partial L}{\partial R} = 2(R-a) + 2u(R-X) = 0
\end{equation}
and get that $R = \frac{a+uX}{1+u}$.
We can also get the complementary slackness that
\begin{equation} \small
    u[(R-X)^T(R-X) - \epsilon_X] = 0
\end{equation}
If $u>0$, we need to have $R = \frac{a+uX}{1+u}$ and $(R-X)^T(R-X) = \epsilon_X$.
By reformulating, we can have $u = \sqrt{\frac{||a-X||_2^2}{\epsilon_X}} -1$.
If $u=0$ and $||a-X||_2^2 \leq \epsilon_X$, we have $R = a$.

\vspace{3pt}
\noindent \textbf{Topology Update.}
In topology update, we aim to minimize the following function by finding $S_{i}$
\begin{equation} \small
    -\ell(S_i, \Tilde{X}_i^{(k+1)}) + I_{C_{Si}}(S_i)
\end{equation}
Similar to feature update, we can first use the gradient descent method to access the gradient $g_{S_i}^{(k)}$ on $S_i$ and then use the projection method to generate the perturbed topology in the feasible set $C_{S_i}$
\begin{equation} \small \label{eq:topologyUpdate}
\begin{split}
    S_i^{(k+1)} & = \Pi_{C_{S_i}}[S_i^{(k)} - \eta_t \cdot g_{S_i}^{(k)}] \\
    g_{S_i}^{(k)} & = -\frac{\partial}{\partial S_i} \ell(S_i, \Tilde{X}_i^{(k+1)}) \\
\end{split}
\end{equation}
where $\eta_t$ is the learning rate.
With the Langrangian function and KKT condition, we can derive the closed-form formula to analytically project $g_{S_i}^k$ to the feasible set $C_{S_i}$. Due to the similarity with proof for Proposition 1 and page limits, we leave the detailed proof to appendix.

\noindent \textbf{Proposition 2.}
 Given $C_{S_i} = \{S_i  \;| \; S_i \in [0,1]^{\frac{N}{M}\times N}, 1^TS_i \leq \epsilon_i \}$, the projection of $a$ to $C_{S_i}$ is
\begin{equation} \small
    \Pi_{C_{S_i}}(a) = \begin{cases}
                P_{[0,1]}(a-u1) \;\; \text{if $u>0$ and $1^TP_{[0,1]}(a-u1) = \epsilon_i$} \\
                P_{[0,1]}(a) \;\; \text{if $1^TP_{[0,1]}(a) \leq \epsilon_i$}
            \end{cases}
\end{equation}
where $P_{[0,1]}(x) = x, if \; x\in [0,1];  0, if \; x<0; 1, if\; x>1.$


\setlength{\textfloatsep}{0pt}
\begin{algorithm}[t] \small
    \caption{SAG to solve Problem~\ref{eq:optimization4}.}
    \label{alg:summary}
    \SetAlgoLined
    \textbf{Input:} Given A, X, fixed GNN weights $W$, learning rate $\eta_t$, epoch number $K$, and partition number $M$;
    
    \textbf{Initialize:}
    $S_i = A_i$, $X_i=X$
    
    \For{$k = 1,2,...,K$}{
        \For{$i = 1,2,...,M$}{
            \textbf{Feature Update on $\Tilde{X}_i$:} 
            
            $\Tilde{X}_i^{k+1} = \Pi_{C_{Xi}}[\Tilde{X}_i^k - \eta_k \cdot g_{X_i}^k]$ with Eq \ref{eq:featureUpdate}.

            \textbf{Topology Update on $S_i$:} 
            
            $S_i^{k+1} = \Pi_{C_{S_i}}[S_i^k - \eta_t \cdot g_{S_i}^k]$ with Eq \ref{eq:topologyUpdate}.

            \textbf{Dual Update on $\mu_i$:} 
            
            $\mu_i^{k+1} = \mu_I^k + \rho \cdot (\Tilde{X}_i^{k+1}-\Tilde{X}_{i+1}^{k+1})$ with Eq \ref{eq:ADMMForm}.        
        }
    }
    Sample and generate the final perturbed matrix.
\end{algorithm}

\vspace{4pt}
\noindent \textbf{Optimization and Complexity Analysis.}
We summarize our SAG in Algorithm \ref{alg:summary}.
There are two optimization loops.
In the inner loop, we iterate through $M$ graph partitions and update the corresponding partitioned graph perturbation $S_i$ and feature perturbation $\Tilde{X}_i$.
At each iteration, only a single graph partition $S_i$ needs to be considered and large memory is saved for not considering the other $M-1$ partitions.
In the outer loop, we repeat the optimization for $K$ (=200 by default) iterations for the algorithm to converge.

After $K$ iterations, $\Tilde{X}_i$ will be identical to each other and can be directly used as the feature perturbation $\Tilde{X}$.
Recalling that $S_i$ is a probability whether an edge needs to be perturbed for attack, we use Bernoulli distribution to sample a $0\text{-}1$-valued edge between each pair of nodes.
We repeat this sample procedure for $20$ times and select the one with minimal loss for the final perturbed topology $S \in \mathcal{R}^{N \times N}$.

The memory complexity of \Mname~is 
\begin{equation} \small
    O(\frac{N}{M} \cdot  N + N\cdot D)
\end{equation}
for storing graph partition $S_i$ and feature perturbation $\Tilde{X}_i$, respectively, since we only need to optimize one graph partition at each iteration.
We store only the current graph partition in the limited GPU memory (around 10 GB) and offload the remaining graph partitions to the large host memory (more than 50 GB).
Note that the same implementation strategy cannot be applied to \cite{xu2019topology} which requires attacking the whole graph $S \in \mathcal{R}^{N\times N}$ at each iteration.







\section{Evaluation}
In this section, we evaluate SAG on five datasets and compare with three attack algorithms to show its effectiveness.

\noindent\textbf{Datasets.}
In this experiment, we select various datasets to cover the vast majority of the GNN inputs, including typical datasets (\textit{Citeseer}, \textit{Cora}, \textit{Pubmed}, and \textit{Amazon-Computer/Photo}) used by many GNN papers~\cite{GCNConv, GINConv, SageConv}. 
Details of these datasets are listed in Table~\ref{tab:dataset}. 

\vspace{3pt}
\noindent\textbf{Baselines.}
To evaluate the effectiveness of SAG, we compare it with the state-of-the-art attack methods by using the adversarial attack repository DeepRobust~\footnote{\url{https://github.com/DSE-MSU/DeepRobust.git}}. 
\begin{itemize}
    \item \textbf{FGSM} \cite{FGSM} is a gradient-based adversial attack that generates adversarial examples by perturbing the input features.
    While it is originally designed for attacking CNNs, it can be easily adapted to attack GNNs by perturbing node features. 
    \item \textbf{PGD} \cite{xu2019topology} is another gradient-based attack method tailored for discrete graph data.
    The reason to select PGD for comparison is that it provides fast attack on discrete graph data by leveraging an optimization-based approach.
    \item \textbf{Nettack} \cite{zugner2018adversarial} is the most popular attack algorithm on graph data by incorporating both edge and feature attacks.
    We select Nettack for comparison because it serves as a strong baseline for SAG on the joint attack.
\end{itemize}

\noindent\textbf{Models.} \underline{\textit{Graph Convolutional Network}} (\textbf{GCN})~\cite{GCNConv} is one of the most popular GNN architectures. It has been widely adopted in node classification, graph classification, and link prediction tasks.
Besides, it is also the key backbone network for many other GNNs, such as GraphSage~\cite{SageConv}, and differentiable pooling (Diffpool)~\cite{diffpool}. We use the setting of hidden dimension size = 16 for each layer of GCN. \underline{\textit{Graph Attention Network}} (\textbf{GAT})~\cite{GINConv}, another typical category of GNN, aims to distinguish the graph-structure that cannot be identified by GCN. 
GAT differs from GCN in its aggregation function, which assigns different weights for different nodes during the aggregation. We use the setting of 8 hidden dimension and 8 attention heads for each layer of GAT.

\vspace{3pt}
\noindent \textbf{Platforms.}
We implement SAG based on PyTorch Geometric~\cite{PyG}. 
We evaluate SAG on Dell T7910 (Ubuntu 18.04) with Intel Xeon CPU E5-2603, 64 GB host memory, and an NVIDIA 1080Ti GPU with 12 GB memory.

\begin{table}[t] \small
    \caption{Datasets for Evaluation.}
    \label{tab:dataset}
    \vspace{-3pt}
    \centering
     \begin{tabular}{ l c c c c }
    \Xhline{2\arrayrulewidth}
    \textbf{Dataset} & \textbf{\#Vertex} & \textbf{\#Edge} & \textbf{\#Dim} & \textbf{{\#Class}}\\
    \Xhline{2\arrayrulewidth}
    Cora	    & 2,708     & 10,858	& 1,433 & 7      \\
    Citeseer    & 3,327	    & 9,464	    & 3,703 & 6      \\
    Amazon-Photo & 7,487 & 119,043 & 745 & 8 \\
    Amazon-Computer & 13,381    & 245,778	    & 767  & 10 \\
    Pubmed	    & 19,717	& 88,676	& 500  & 3      \\
    \Xhline{2\arrayrulewidth}
    \end{tabular}
    \vspace{3pt}
\end{table}

\begin{table*}[t] \small
    \centering
    \caption{Evaluation of \Mname~with existing adversarial attacks.}
        \vspace{-5pt}
    \scalebox{0.93}{
    \begin{threeparttable}
    \begin{tabular}{|c||c|c|c|c|c|}
    \hline
        \textbf{Dataset} & \textbf{Method}  & \textbf{Time (min) } &\textbf{Mem. (GB)} & \textbf{Evasive Acc. (\%)} & \textbf{Poisoning Acc. (\%)}\\
    \hline
    \hline
        \multirow{5}{*}{Cora}&Clean & 0 & 0 & 79.63 & 79.63 \\
        & FGSM & 0.05 & 0.64 & 70.52 & 78.87 \\
        & PGD &  0.25 & 1.03 & 75.70 & 70.77 \\
        & Nettack &  14.75 & 0.68 & 69.62 & 71.35 \\
        & \textbf{SAG} &  0.48 & 0.70  & \textbf{68.06} & \textbf{67.15} \\
    \hline
    \hline
        \multirow{5}{*}{Citeseer}&Clean & 0 & 0 & 71.80 & 71.80 \\
        & FGSM & 0.03 & 0.41 & 67.00 & 71.70 \\
        & PGD &  0.18 & 0.97 & 69.08 & 67.30 \\
        & Nettack & 12.63 & 0.60 & 62.91 & 67.54 \\
        & \textbf{SAG} & 0.34 & 0.82 & \textbf{62.62} &  \textbf{63.68}\\
    \hline
    \hline
        \multirow{5}{*}{\thead{Amazon\\Photo}}&Clean & 0 & 0 & 93.11 & 93.11 \\
        & FGSM & 0.05 & 1.17 & 89.13 & 91.37 \\
        & PGD & 4.21 & 3.69  &  83.02 &  82.77\\
        & Nettack & 1560 & 1.84 & 87.95 & 89.25\\
        & \textbf{SAG} & 6.13 & 1.24 & \textbf{80.52} & \textbf{80.12} \\
    \hline
    \hline
        \multirow{5}{*}{\thead{Amazon\\Computer}}&Clean & 0 & 0 & 89.29 & 89.29 \\
        & FGSM & 3.28 & 2.67 & 85.62 & 88.03 \\
        & PGD &  17.32 & 10.59  & 77.22 & 77.93 \\
        & Nettack & 2578 & 4.35 & 82.41 & 85.33 \\
        & \textbf{SAG} & 18.78 & 2.91 & \textbf{74.56} & \textbf{76.31} \\
    \hline
    \hline
        \multirow{5}{*}{Pubmed}&Clean & 0 & 0 & 77.53 & 77.53 \\
        & FGSM &  9.83 & 3.54 & 72.93 & 74.25 \\
        & PGD & - & \underline{\textit{OOM}} & - & - \\
        & Nettack & 3108 & 8.17 & 67.72 & 76.25  \\
        & \textbf{SAG} & 47.62 & 3.63 & \textbf{62.71} & \textbf{36.27} \\
    \hline
    \end{tabular}
     \begin{tablenotes}
      \small
      \item[1] \textbf{Note that ``-'' means such parameter is not applicable to the given setting.}
      \item[2] \textbf{OOM refers to ``Out of Memory''.}
    \end{tablenotes}
    \end{threeparttable}
    }
    \vspace{-15pt}
    \label{tab: Overall Quantization Performance (Accuracy, Average Bits, and Memory Saving}
\end{table*}

\vspace{3pt}
\noindent\textbf{Metrics.}
We evaluate \Mname~with six metrics -- \textit{evasive accuracy}, \textit{poisoning accuracy}, \textit{topology ratio}, \textit{feature ratio}, \textit{memory consumption}, and \textit{running time}.
Following the common setting~\cite{zugner2018adversarial, xu2019topology}, we report the \textbf{evasive accuracy} by assuming the GNN model is fixed and targeting the test data.
We report the \textbf{poisoning accuracy} by targeting the training data and perform the model training phase after the attack.
The \textbf{topology ratio} (\%) is computed as the number of attacked edges over the number of existing edges in the clean graph dataset.
The \textbf{feature ratio} (\%) is reported as the $L_2$ norm of perturbed features over the $L_2$ norm of the original features in the clean graph dataset.
To measure the memory, we utilize NVProf to query the runtime GPU memory consumption at a pre-selected frequency of $100$ ms.
To measure the time, we leverage the software timer from python $time$ library.
For a fair comparison, all gradient-based approaches are conducted for $200$ iterations.



\vspace{-5pt}
\subsection{Overall Performance}
Table ~\ref{tab: Overall Quantization Performance (Accuracy, Average Bits, and Memory Saving} shows the overall performance comparison between \Mname~and existing adversarial attacks under the same setting of topology ratio and feature ratio.
Following the most common setting used by many previous papers~\cite{zugner2018adversarial, xu2019topology}, we select the same topology attack ratio of $5\%$ and feature attack ratio of $2\%$, and leave the study on diverse ratios to the ablation study.
On PGD and FGSM, we only attack the topology and features, respectively, due to their limits in attacking capability.
In SAG, we stick to \textit{M}=2 and will exhibit the impact of \textit{M} in the ablation study.
We observe that \Mname~consistently outperforms the state-of-the-art attack approach, such as FGSM, PGD, and Nettack on evasive attack (up to $14.82\%$ accuracy drop) and poisoning attack (up to $41.26\%$ accuracy drop) across different datasets.
On Pubmed dataset, \Mname~achieves $14.82\%$ accuracy drop in evasive attack and $41.26\%$ in poisoning attack.
The major reason for such success is that SAG enables the gradient-based joint optimization on both the features and topology while incorporating global reasoning on the interaction between attacking different nodes.
By contrast, FGSM and PGD attack only the feature or topology, and Nettack considers only one edge at each time, failing to reason the global interaction across edges and nodes. 

\begin{figure*}
    \begin{subfigure}{0.32\textwidth}
        \centering
        \includegraphics[width=\linewidth, height=3.0cm, trim=0 0.1cm 0 0]{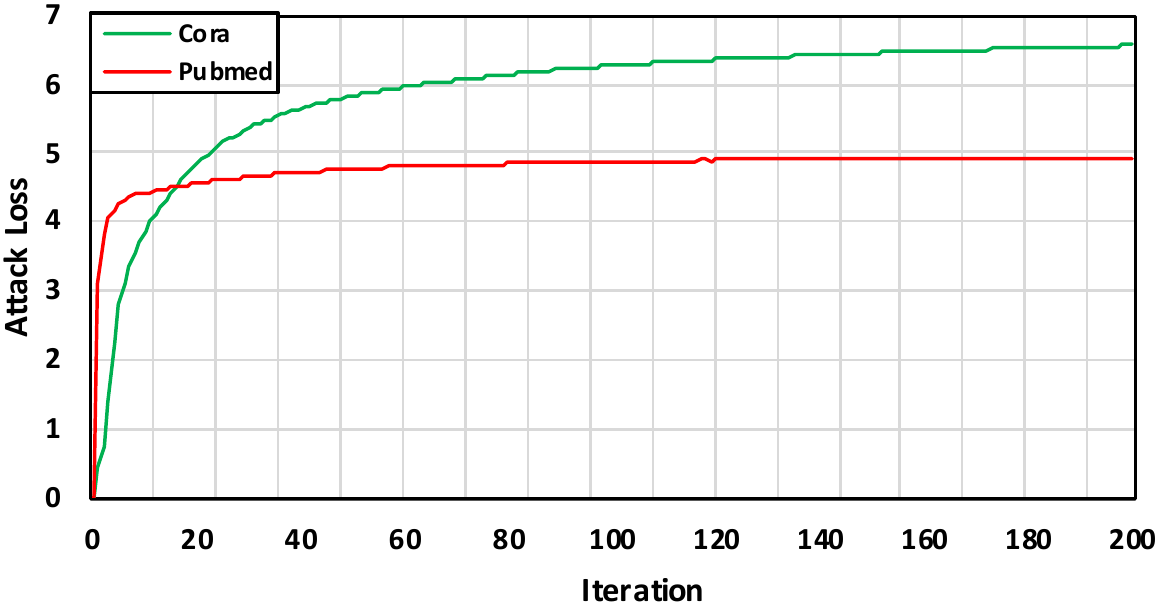}
        \caption{Attack Loss}
        \label{fig:attackLoss}
    \end{subfigure}
    \begin{subfigure}{0.32\textwidth}
    \centering
    \includegraphics[width=\linewidth, height=3cm, trim=0 0.1cm 0 0]{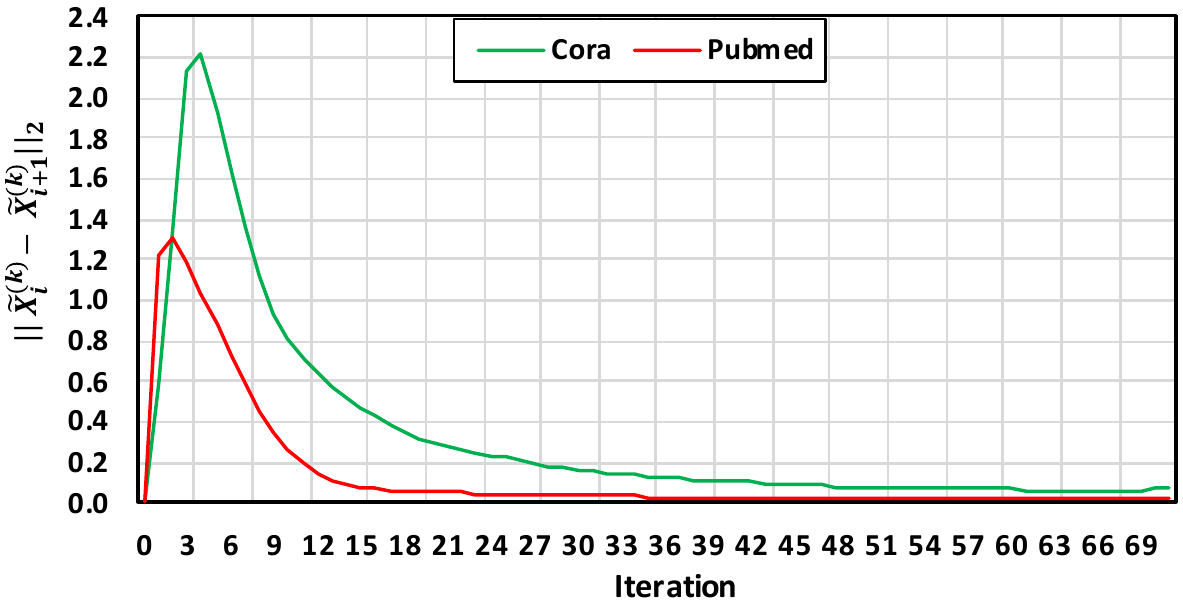}
    \caption{\small $||\Tilde{X}_i^{(k)} - \Tilde{X}_{i+1}^{(k)}||_2$}
    \label{fig:convergenceBehavior-1}
    \end{subfigure}
    \begin{subfigure}{0.32\textwidth}
    \centering
    \includegraphics[width=\linewidth, height=3.1cm, trim=0 0.1cm 0 0]{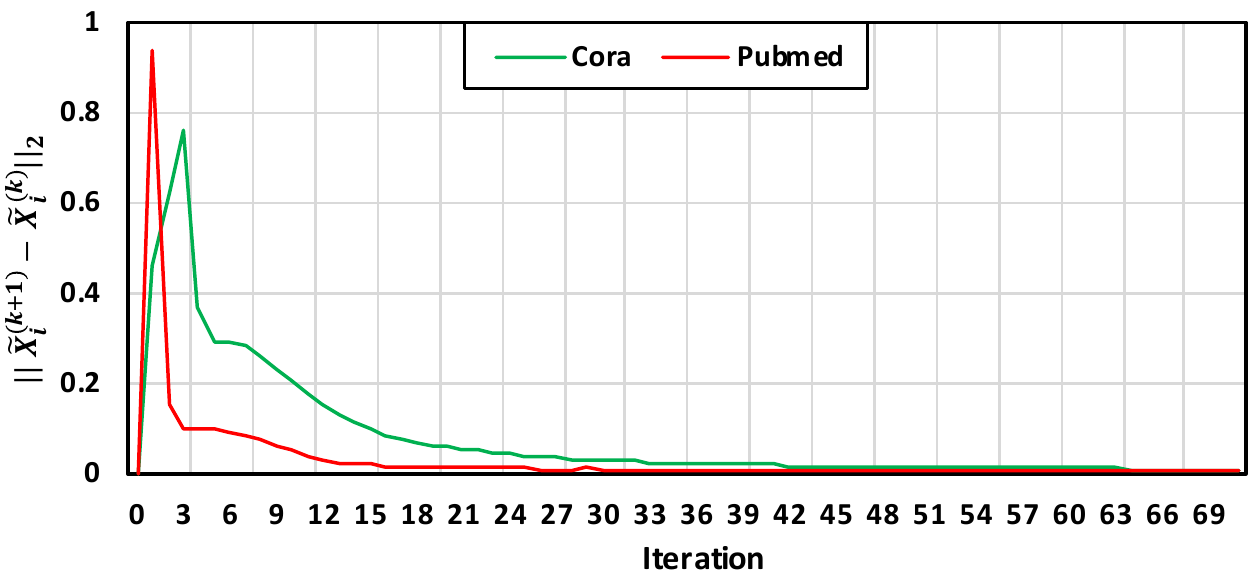}
    \caption{$||\Tilde{X}_i^{(k+1)} - \Tilde{X}_i^{(k)}||_2$}
    \label{fig:convergenceBehavior-2}
    \end{subfigure}
    \caption{Convergence Behavior of \Mname~on \textbf{Cora} and \textbf{Pubmed}.}
    \vspace{-15pt}
    \label{fig:convergenceBehavior}
\end{figure*}

Across different datasets and settings, we notice that Nettack always comes with the highest time cost.
The reason is that, at each iteration, it selects only one edge or feature to attack by examining all edges and node features, and repeats the procedure until reaching the topology ratio and the feature ratio.
Moreover, we observe that PGD usually has the highest memory consumption, since it requires a floating-point $N\times N$ matrix to store the edge gradients between each pair of nodes, where $N$ is the number of nodes.
For Pubmed, a single matrix of this shape requires at least $1.5$ GB to store and PGD processes the whole matrix at each iteration, instead of processing only a small partition as the case in SAG.
Besides, in the time and memory comparison among these implementations,  we notice the significant strength of \Mname, which achieves up to $254\times$ speedup and $3.6\times$ memory reduction.
This is largely due to our effective algorithmic and implementation optimizations that can reduce the runtime complexity meanwhile amortizing the memory overhead to great extent.

\subsection{Ablation Studies}
In this ablation study, we will focus on two representative datasets -- Cora and Pubmed -- for the performance of SAG on a small graph dataset and a large graph dataset.

\textbf{ADMM Convergence Behavior}
We show the ADMM convergence behavior in Figure ~\ref{fig:convergenceBehavior}.
Here, we adopt the same number of epochs as $T=200$ and show only the first $70$ epochs in Figure \ref{fig:convergenceBehavior-1} and Figure \ref{fig:convergenceBehavior-2} since SAG converges fast on $||\Tilde{X}_i^{(k)} - \Tilde{X}_{i+1}^{(k)}||_2$ and $||\Tilde{X}_i^{(k+1)} - \Tilde{X}_i^{(k)}||_2$.
We only present the result for $i=1$ since various $i$'s show similar results.
Overall, SAG converges gracefully as epoch increases, demonstrating the effectiveness of our method.
In Figure~\ref{fig:attackLoss}, with the increase of the epoch number, SAG gradually increases the attack loss by perturbing the features and the topology.
In Figure \ref{fig:convergenceBehavior-1}, $||\Tilde{X}_i^{(k)} - \Tilde{X}_{i+1}^{(k)}||_2$ starts from 0 since we initialize both $\Tilde{X}_{i}$ and $\Tilde{X}_{i+1}$ to the node features on clean graph dataset.
Compared across different datasets, Pubmed converges faster than Cora since Pubmed contains much smaller feature dimension than Cora.




\textbf{Topology and Feature Perturbation}
For poisoning attack on Cora dataset (Table~\ref{tab:ablationCora}), the increase of feature perturbation and topology perturbation would both lead to the accuracy drop compared with the original clean data.
Besides, we observe that, at the same level of topology perturbation, $8\%$ feature perturbation can lead to $3.6\%$ extra accuracy drop on average.
On Pubmed dataset, we observe that $8\%$ feature perturbation can lead to $9.4\%$ extra accuracy drop, averaged over various topology perturbation ratio.
These results show the benefit of attacking both topology and features.


\begin{table}[t] \small
    \centering
    \caption{Accuracy (\%) of \Mname~under the diverse ratio of perturbed edges and feature attacks on \textbf{Cora} and \textbf{Pubmed}.}
    \vspace{-5pt}
    \scalebox{0.97}{
    \begin{tabular}{c|c|c|c|c|c|c}
    \Xhline{2\arrayrulewidth}
    \textbf{Cora} & \multicolumn{6}{c}{\textbf{Feature Perturbation (\%)}}  \\
    \hline
        \multirow{6}{*}{\shortstack{\textbf{Topology} \\ \textbf{Perturbation} \\ \textbf{(\%)}}}&  & \textbf{0} &  \textbf{1}  & \textbf{2} & \textbf{4} & \textbf{8} \\
        & \textbf{0}  & 79.63 & 79.25 & 78.91 & 78.84 & 78.73 \\
        & \textbf{5}  & 70.47 & 69.11 & 68.71 & 66.52 & 65.59 \\
        & \textbf{10} & 58.63 & 57.29 & 56.34 & 55.23 & 53.77 \\
        & \textbf{15} & 50.15 & 49.60 & 47.53 & 47.04 & 46.73 \\
        & \textbf{20} & 45.37 & 43.46 & 43.06 & 42.71 & 41.65 \\
    \Xhline{2\arrayrulewidth}
    \textbf{Pubmed} & \multicolumn{6}{c}{\textbf{Feature Perturbation (\%)}}  \\
    \hline
        \multirow{6}{*}{\shortstack{\textbf{Topology} \\ \textbf{Perturbation} \\ \textbf{(\%)}}}&  & \textbf{0} & \textbf{1} & \textbf{2} & \textbf{4} & \textbf{8} \\
        & \textbf{0}  & 77.53 & 75.34 & 73.91 & 72.55 & 68.65 \\
        & \textbf{5}  & 49.19 & 39.14 & 36.27 & 31.65 & 28.53 \\
        & \textbf{10} & 33.24 & 28.57 & 27.81 & 26.35 & 25.30 \\
        & \textbf{15} & 28.04 & 26.05 & 23.59 & 23.04 & 22.34 \\
        & \textbf{20} & 23.95 & 21.62 & 21.09 & 20.71 & 20.38 \\
    \hline
    \end{tabular}
    }
    \label{tab:ablationCora}
    \vspace{-10pt}
\end{table}

\begin{table}[t] \small
    \centering 
    \caption{Impact of \textit{M} for Poisoning Attack on \textbf{Pubmed}.}
    \vspace{-5pt}
    \begin{tabular}{c|c|c|c}
        \Xhline{2\arrayrulewidth}
        \multirow{2}{*}{\textbf{M}} & \textbf{Time} & \textbf{Memory} & \textbf{Accuracy Drop} \\
        & \textbf{(min)}& \textbf{(GB)} & \textbf{(\%)}\\
        \hline
        \textbf{1} & 33  & 5.72 & 35.37 \\
        \textbf{2} & 42 & 3.63 & 36.27\\
        \textbf{4} & 62 & 2.75 & 36.18 \\
        \textbf{8} & 95  & 2.21 & 35.97 \\
        \Xhline{2\arrayrulewidth}
    \end{tabular}
    \label{tab:ablationK}
    \vspace{5pt}
\end{table}

\begin{table}[t]
    \centering
    \caption{SAG Transferability for \textbf{Poisoning} Attack.}
    \vspace{-5pt}
    \begin{threeparttable}
    \begin{tabular}{c|c|c}
        \Xhline{2\arrayrulewidth}
         & \textbf{Cora (\%)} & \textbf{Pubmed (\%)} \\
        \hline
        \textbf{1-layer GCN} & 0.67 (0.80) & 0.35 (0.77)\\
        \textbf{2-layer GCN} & 0.78 (0.83) & 0.80 (0.85) \\
        \textbf{4-layer GCN} & 0.74 (0.81) & 0.76 (0.84)\\
        \hline
        \textbf{1-layer GAT} & 0.74 (0.82) & 0.37 (0.79)\\
        \textbf{2-layer GAT} & 0.77 (0.83) & 0.78 (0.85) \\
        \textbf{4-layer GAT} & 0.75 (0.80) & 0.73 (0.82)\\
        \Xhline{2\arrayrulewidth}
    \end{tabular}
    \begin{tablenotes}
      \small
      \item[1] Data Format: \textbf{attacked data acc. (clean data acc.).}
    \end{tablenotes}
    \end{threeparttable}
    \label{tab:SAG Transferability for Poisoning Attack}
    \vspace{3pt}
\end{table}

\textbf{M-value Impact} We also evaluate SAG for poisoning attack on Pubmed to show the impact of the hyperparameter \textit{M} (\textit{i.e.}, the number of graph partitions) on memory saving.
As shown in Table~\ref{tab:ablationK}, with the increase of the \textit{M} value, the memory size reduction becomes significant, since splitting the graph into \textit{M} partitions and attacking individual partitions at each time essentially reduce the memory requirement.
We also observe similar accuracy drop under different \textit{M} since SAG converges gracefully and hits similar optimal points for diverse \textit{M}.
Meanwhile, we also observe that the increase of value \textit{M} also brings the runtime overhead in terms of time cost, for example, \textit{M}=8 setting is 33 minutes slower than \textit{M}=4 setting. 
This slowdown happens since we need to attack individual split graphs at each time, leading to a small portion of system overhead on memory access.
This also leads to a tradeoff among these factors when selecting the value of \textit{M}.
We also observe that the memory consumption does not decrease linearly as \textit{K} increases.
The main reason is that, as \textit{M} reduces, memory consumption from other sources (\textit{e.g.}, loading PyTorch framework and the features) becomes the dominant component.







\vspace{3pt}
\textbf{Transferability} To demonstrate the transferability of \Mname, we further evaluate our attacked graphs (Cora and Pubmed) on GCNs (with 1, 2, and 4 layers) and GATs (with 1, 2, and 4 layers), respectively.
We generate adversarial examples on a 1-layer GCN model and conduct poisoning attack on other models by targeting the training data and training these models on the perturbed data.
As shown in Table~\ref{tab:SAG Transferability for Poisoning Attack}, \Mname~can effectively maximize the accuracy drop on Cora (up to 13\%) and Pubmed (up to 42\%).
The major reason for such success in launching the poisoning attack is that the adversarial attack on 1-layer GCN effectively captures the intrinsic property on the graph data that is agnostic to the models.
We want to stress that, even on models with different layers (\textit{i.e.}, 2 and 4), the poisoned graph data can still achieve $8\%$ and $9\%$ accuracy drop on Cora and Pubmed, respectively.
These results demonstrate the transferability of \Mname~towards models with diverse architectures and the number of layers.

\vspace{-5pt}
\section{Conclusion}
This work focuses on GNN robustness by giving an in-depth understanding of GNN's ``weakness''. We propose \Mname, the first scalable adversarial attack method with Alternating Direction Method of Multipliers (ADMM), which can successfully overcome the limitations of the previous solutions. Extensive experiments further highlight \Mname's advantage of reducing the computation and memory overhead over the existing approaches.

\newpage

\bibliography{reference}

\newpage

\todo{ }
\newpage

\section{Proof of Proposition 2}
In this section, we provide the proof for Proposition 2.
Similar to the proof in Proposition1 and existing works on projection \cite{DBLP:conf/aaai/LengDLZJ18,ADMM-Pruning,xu2019topology}, we utilize the Langrangian function and the KKT contition to derive a closed-form formula to project a given input towards the feasible set $C_{S_i}$.


\noindent \textbf{Proposition 2.}
 Given $C_{S_i} = \{S_i  \;| \; S_i \in [0,1]^{\frac{N}{M}\times N}, 1^TS_i \leq \epsilon_i \}$, the projection of $a$ to $C_{S_i}$ is
\begin{equation} \small
    \Pi_{C_{S_i}}(a) = \begin{cases}
                P_{[0,1]}(a-u1) \;\; \text{if $u>0$ and $1^TP_{[0,1]}(a-u1) = \epsilon_i$} \\
                P_{[0,1]}(a) \;\; \text{if $1^TP_{[0,1]}(a) \leq \epsilon_i$}
            \end{cases}
\end{equation}
where $P_{[0,1]}(x) = x, if \; x\in [0,1];  0, if \; x<0; 1, if\; x>1.$

\noindent \textbf{Proof: }
We first transform the projection problem $\Pi_{C_{S_i}}$ into an optimization problem
\begin{equation} \small
\begin{split}
    \min_R \;\;\;\; & \frac{1}{2}||R-a||_2^2 \\
    \text{s.t.} \;\;\;\; & R \in [0,1]^{\frac{N}{M} \times N} \\
    & 1^tR\leq \epsilon_i
\end{split}
\end{equation}
Then, we can derive its Langrangian function as
\begin{equation} \small
    L(R,u) = \frac{1}{2} ||R-a||_2^2 + I_{[0,1]}(R) + u(1^TR-\epsilon_i)
\end{equation}
where $u\leq 0$ is the dual variable.
Here, $I_{[0,1]}(R) = 0 \text{ if } R \in [0,1]^{\frac{N}{M}\times N}; =\infty \text{  otherwise.}$
Using the KKT condition, we have the stationary condition that
\begin{equation} \small
    \frac{\partial L}{\partial R} = (R-a) + u1 + \frac{\partial}{\partial R}I_{[0,1]}(R) = 0
\end{equation}
Here, $\frac{\partial}{\partial R}I_{[0,1]}(R) = 0 \text{ if } R \in [0,1]^{\frac{N}{M}\times N}; =\infty \text{  otherwise}$
We have $R = P_{[0,1]}(a-u1)$, where $P_{[0,1]}(x) = x, if \; x\in [0,1]; = 0, if \; x<0; = 1, if\; x>1$, and $P_{[0,1]}(x)$ is element-wisely applied on $(a-u1)$.
Using the KKT condition, we also have the complementary slackness
\begin{equation} \small
    u(1^TR-\epsilon_i) = 0
\end{equation}
If $u=0$, we need to have $R = P_{[0,1]}(a)$.
If $u>0$, we need to have $R = P_{[0,1]}(a-u1)$ and $1^TR-\epsilon_i = 0$.
In other words, we have $1^TP_{[0,1]}(a-u1) = \epsilon_i$, where $u$ is a scalar variable and can be solved with the bisection method \cite{ADMM}.

\end{document}